\documentclass[10pt,twocolumn]{ICCAS2023}

\DeclareFontFamily{U}{mathc}{}
\DeclareFontShape{U}{mathc}{m}{it}%
{<->s*[1.03] mathc10}{}

\DeclareMathAlphabet{\mathscr}{U}{mathc}{m}{it}

\usepackage{diagbox}
\usepackage{amsmath, amssymb}

\usepackage{ulem} 

\begin{document}

\title{Dimensionally Homogeneous Jacobian using Extended Selection Matrix for Performance Evaluation and Optimization of Parallel Manipulators }

\author{Hassen Nigatu${}^{1}$ and Doik Kim${}^{1*}$ }

\affils{ ${}^{1}$Center for Intelligent \& Interactive Robotics Research, Korea Institute of Science and Technology, \\
Seoul, 02792, Korea.  Email: (614503@kist.re.kr),
${}$  (doikkim@kist.re.kr) {\small${}^{*}$ Corresponding author}}

\thanks{ \noindent
   This work was supported by Korea Institute of Science and Technology (KIST), under Grant 2E32302.
  }

\abstract{
   This paper proposes a new methodology for deriving a point-based dimensionally homogeneous Jacobian, intended for performance evaluation and optimization of parallel manipulators with mixed degrees of freedom. Optimal manipulator often rely on performance indices obtained from the Jacobian matrix. However, when manipulators exhibit mixed translational and rotational freedoms, the conventional Jacobian's inconsistency of units lead to unbalanced optimal result.
   Addressing this issue, a point-based dimensionally homogeneous Jacobian has appeared as a prominent solution. However, existing point-based approaches for formulating dimensionally homogeneous Jacobian are applicable to a limited variety of parallel manipulators. Moreover, they are complicated and less intuitive. 
   This paper introduces an extended selection matrix that combines component velocities from different points to describe the entire motion of  moving plate. This proposed approach enables us to formulate an intuitive point-based, dimensionally homogeneous Jacobian, which can be applied to a wide variety of constrained parallel manipulators.
   To prove the validity of proposed method, a numerical example is provided utilizing a four-degree-of-freedom parallel manipulator. 
     }
\keywords{
    Dimensionally homogeneous Jacobian, Selection matrix, Inverse Jacobian, Parallel manipulator, Performance indices
}

\maketitle


\section{Introduction} \label{sec:intro}

	Performance evaluation and obtaining optimized architectural parameters are a vital step in the design of parallel manipulators (PMs), as these significantly influences the effectiveness and accuracy of a robot's movements. The challenge lies in performing the tasks when the manipulators degree of freedom (DoFs) is a combination of rotational and translational types. This is primarily due to the inconsistency in the units or dimensions of the Jacobian, a factor that significantly affects the performance measuring indices of parallel manipulators \cite{Gosselin1992,Merlet2007}.
	
	Several approaches have been suggested to address this problem \cite{Yang2022,Rosyid2020}, and among them, Jacobian-based methods have been widely used. This popularity can be attributed to their capability to effectively translate the inherent mapping process from joint velocities to end-effector velocities, providing a good intuitive framework \cite{Gosselin1992,Kim2003,Pond2006}. There are also variety of Jacobian-based approaches of homogenizing the units of the Jacobian matrix \cite{Rosyid2020,Nigatu2021thesis}. Among these approaches, point-based approach is more intuitive \cite{Pond2006,Nigatu2023}. Despite its advantage, the firsly introduced point-based dimensionally homogeneous Jacobian (DHJ) formulations comprise dependent motions in their entries, resulting in a condition number with unclear physical meaning and potential erroneous results \cite{Pond2006}. 
	
	In response to this problem, Pond et al. \cite{Pond2006} proposed a method to eliminate the undesired dependent motions from the system. However, the method is quite complicated to comprehend and involved a tedious derivative procedures which leads to higher computation cost \cite{Nigatu2023}. To overcome this issue, the selection matrix with the shifting property and conventional Jacobian is used to formulate a point-based dimensionally homogeneous Jacobian matrix \cite{Nigatu2023}. However, this previous paper by the authors focused on a specific scenario where each component's velocity encompassed the desired motion of the moving plate, such as 1T2R PMs with $TzRxRy$ type of motion.\\	
	Considering the aforementioned limitations, this paper formulates an $f \times f$, with $f$ representing the DoF of the mechanism, point-based DHJ matrix using an extended selection matrix. This Jacobian matrix maps the platform's nominal linear velocity to the joint rate. Here, nominal linear velocity refers to the velocity obtained by combining component velocities from different points, which can represent the entire motion of the moving plate. This approach integrates the extended selection matrix, the linear velocity of points on the moving plate and the manipulator's conventional Jacobian, resulting in a square DHJ. 	
	The dimensionally homogeneity of the resulting Jacobian is analytically proven. To validate the correctness of the proposed method, a numerical comparison is carried out using a four-degree-of-freedom parallel manipulator as an example. First, the distribution of the condition number is evaluated across the manipulator's rotational workspace, highlighting the disparity in the condition number values of the conventional and dimensionally homogeneous Jacobian. Then, the unit of geometric parameters are changed from millimeters to meters, and the condition number is reassessed to determine if it is invariant under unit changes. 
\section{Formulation of the Dimensionally Homogeneous Jacobian} \label{sec:FDHJ}
    The derivation method of DHJ involves the following steps. First, the screw-based constraint embedded inverse Jacobian of the manipulator is formulated and inverted to get the constraint compatible forward relation. Then, points that might adequately represent the motion of the moving plate are chosen and related to the Cartesian velocity. Next, the extended selection matrix is derived and applied to the points' linear velocity. This will combine components from different points to effectively describe the moving plate's motion, while also eliminating unwanted or dependent components from the equation. The resulting velocity is termed as the nominal linear velocity. Finally, the nominal linear velocity of the moving plate and the forward velocity equation are related with an $f \times f$ dimensionally homogeneous Jacobian matrix.

	\subsection{Constraint-Embedded Velocity Relation} \label{subsec:CEVR}
	The screw-based Jacobian of the manipulator can be analytically obtained using the method introduced in \cite{Kim2003,Kim2002,Nigatu2021,Nigatu2021_1}. Given the task velocity, $\boldsymbol{\dot{\mathscr{x}}}$, of the moving moving plate, the general inverse velocity equation of the parallel manipulator has the following form.

	\begin{equation}
		\begin{bmatrix} \dot{\boldsymbol{q}} \\ \boldsymbol{0} \end{bmatrix}  = \begin{bmatrix} \boldsymbol{G}_{a}^T \\ \boldsymbol{G}_{c}^T  \end{bmatrix} \boldsymbol{\dot{\mathscr{x}}} = \begin{bmatrix} \boldsymbol{G}_{av}^T & \boldsymbol{G}_{aw}^T \\ \boldsymbol{G}_{cv}^T & \boldsymbol{G}_{cw}^T  \end{bmatrix} \begin{bmatrix} \boldsymbol{v} \\ \boldsymbol{\omega} \end{bmatrix} \label{eq:irk_std}
	\end{equation}

	The units of the entries in $\boldsymbol{G}$ in Eq. (\ref{eq:irk_std}), are dependent on the type of actuators employed in the manipulator. This paper focus exclusively on scenarios where the manipulator employs only linear or rotational actuators, not considering situations involving a combination of these actuator types.

	Inverting Eq. (\ref{eq:irk_std}) yields a constraint compatible forward velocity relation as

	\begin{equation}
		\boldsymbol{\dot{\mathscr{x}}} = \boldsymbol{J}\dot{\boldsymbol{q}} = \begin{bmatrix}
									\boldsymbol{J}_a & \boldsymbol{J}_c
							   \end{bmatrix}
							   \begin{bmatrix} \dot{\boldsymbol{q}}_a \\ \boldsymbol{0}  \end{bmatrix} \label{eq:frk_std}
	\end{equation}

     In Eq. (\ref{eq:frk_std}), $\boldsymbol{J} \in {\mathbb{R}}^{6 \times 6}$ is the inverse of $\boldsymbol{G}^T$ and its sub-matrix $\boldsymbol{J}_c$ is related to the constraint. Thus, we can explicitly describe the relation of $\boldsymbol{\dot{\mathscr{x}}}$ and $\dot{\boldsymbol{q}}_a$ as

     \begin{equation}
     	\boldsymbol{\dot{\mathscr{x}}} = \boldsymbol{J}_a\dot{\boldsymbol{q}}_a,  ~\text{where} ~\boldsymbol{J}_a = \begin{bmatrix} \boldsymbol{J}_{a1} \\ \boldsymbol{J}_{a2}   	\end{bmatrix} \label{eq:frk_simp}
     \end{equation}

 	The Cartesian velocity, $\boldsymbol{\dot{\mathscr{x}}} \in {\mathbb{R}}^{6 \times 1}$, in Eq. (\ref{eq:frk_simp}) is constraint compatible.When the manipulator employs linear actuators, $\boldsymbol{J}_{a1}$ is dimensionless, while $\boldsymbol{J}_{a2}$ has a unit of $\frac{1}{\text{length}}$. Conversely, if the manipulator utilizes rotational actuators,  $\boldsymbol{J}_{a1}$ has a unit of length and $\boldsymbol{J}_{a2}$ dimensionless. Considering these distinctions, the point's linear velocity and selection matrix are established to ensure consistency or removal of units in the Jacobian.

	\subsection{Linear Velocity of Points} \label{subsec:LVP}

    According to the well known shifting property \cite{Lipkin2004} in the rigid body kinematics, any points velocity on the moving plate can be related to the Cartesian velocity of the moving plate as

    \begin{equation}
    	\boldsymbol{v}_i = \boldsymbol{v} + \boldsymbol{\omega} \times  \boldsymbol{a}_i \label{eq:pts_vel}
    \end{equation}
	where $\boldsymbol{v}$ and $\boldsymbol{\omega}$ denotes the linear and angular velocity of the moving plate, while $\boldsymbol{a}_i$ corresponds to a constant vector extending from the origin of the Cartesian reference frame to the $i^{th}$ point on the moving plate. Expanding Eq. (\ref{eq:pts_vel}) reveals the motion of the moving plate that each component of $\boldsymbol{v}_i$ encompasses.

	\begin{equation}
		\begin{aligned}
		   {v}_{ix} &= {v}_x + {\omega}_y {a}_{iz} -{\omega}_z {a}_{iy}  \\
		   {v}_{iy} &= {v}_y - {\omega}_x {a}_{iz} +{\omega}_z {a}_{ix}  \\
		   {v}_{iz} &= {v}_z +{\omega}_x {a}_{iy} -{\omega}_y {a}_{ix}
		\end{aligned} \label{eq:vi_comp}
	\end{equation}

	 By distributing these points on the moving plate in a noncollinear manner, it is possible to satisfy the minimum requirement of points needed to fully represent the motion of the moving plate. Theoretically, the translations of three noncollinear points on the moving plate are sufficient to uniquely identify the motion of the body in terms of translation and rotation, but more points may be required depending on the DoF of the mechanism.

	Hence, Eq. (\ref{eq:pts_vel}) can be generalized as
	\begin{equation}
		\begin{aligned}
		\boldsymbol{v}_p =\begin{bmatrix} \boldsymbol{v}_1 \\ \vdots \\ \boldsymbol{v}_i	\end{bmatrix} &= \begin{bmatrix}\boldsymbol{I} & -[\boldsymbol{a}_1]_\times \\ \vdots & \vdots \\ \boldsymbol{I} & -[\boldsymbol{a}_i]_\times \end{bmatrix}  \begin{bmatrix} \boldsymbol{v} \\ \boldsymbol{\omega}\end{bmatrix} \\
		                 &= \boldsymbol{V}_p\boldsymbol{\dot{\mathscr{x}}} \label{eq:vp}
	\end{aligned} 
	\end{equation}
	where $\boldsymbol{V}_p \in {\mathbb{R}}^{3f \times 6} $ maps the moving plate cartesian velocity to the points velocity on the moving plate. Vector $\boldsymbol{v}_i$ in Eq. (\ref{eq:vp}) has three components and hence from $\boldsymbol{v}_p \in {\mathbb{R}}^{3f \times 1} $, we need to determine the components that can appropriately describe the motion of the moving plate via a selection matrix \cite{Nigatu2023} as follows
	\begin{equation}
		\begin{aligned}
		\boldsymbol{S}\boldsymbol{v}_p     & = \boldsymbol{S}\boldsymbol{V}_p\boldsymbol{\dot{\mathscr{x}}},  ~\text{where} ~ \boldsymbol{S}\in {\mathbb{R}}^{f \times 3f} ~\text{is a selection}  \\&  \text{matrix that}  ~\text{extracts the components from} ~ \boldsymbol{v}_p.\\
		              \boldsymbol{v}_{ps}  & = \boldsymbol{V}_{ps}\boldsymbol{\dot{\mathscr{x}}} ~ \text{where}~ \boldsymbol{V}_{ps}\in {\mathbb{R}}^{f \times 6}
	\end{aligned}	\label{eq:vps}
	\end{equation}

 	However, deriving the selection matrix $\boldsymbol{S}$ is not always straightforward. This is because only manipulators whose moving plates exhibit $T_xR_yR_z$, $T_yR_xR_z$ and $T_zR_xR_y$ types of motion can be uniquely represented with the component velocities shown in Eq. (\ref{eq:vi_comp}). For a comprehensive understanding of the establishment of selection matrices for these groups of PMs, readers are encouraged to refer to \cite{Nigatu2023}. PMs falling outside of these categories will need to utilize a combination of components from different points, an approach that is covered in this paper.
    \subsection{Dimensionally Homogeneous Jacobian} \label{subsec:DHJ}
    
   In this paper, we derive the dimensionally homogeneous Jacobian by representing the motion of the moving plate using linear velocity, ensuring uniform units across its entries. However, it is important to note that the linear velocities used here are not merely the component velocities of individual points on the moving plate. Instead, they are a combination of components from various points. This approach is used to encompass all desired motion of the moving plate into a representative velocity equation, which we call it the nominal velocity.  

   To derive the dimensionally homogeneous Jacobian, relations, Eq. (\ref{eq:frk_simp}) and Eq. (\ref{eq:vps})  are combined as follows

    \begin{equation}
   	\begin{aligned}
   		\boldsymbol{v}_{ps} &= \boldsymbol{V}_{ps}\boldsymbol{\dot{\mathscr{x}}} \\
   		&= \boldsymbol{V}_{ps}\boldsymbol{J}_a\dot{\boldsymbol{q}}_a \\
   		&= \boldsymbol{J}_{dh}\dot{\boldsymbol{q}}_a
   	\end{aligned} \label{eq:jdh}
   \end{equation}

   In Eq. (\ref{eq:jdh}),  $\boldsymbol{J}_{dh} \in {\mathbb{R}}^{f \times f}$ is a Jacobian that relates the nominal linear velocity ($\boldsymbol{v}_{ps}$) of the moving plate to the actuated joint rate ($\dot{\boldsymbol{q}}_a \in {\mathbb{R}}^{f \times 1}$). To demonstrate the consistency of units in its entries, we considered the following two generic cases.

   \textit{Case 1: PMs with linear actuators.} In this case, $\dot{\boldsymbol{q}}_a$ has unit of $\frac{\text{length}}{\text{time}}$ while $\boldsymbol{S}$ is dimensionless. Referring to Eq. (\ref{eq:vp}), we can observe that the first term is dimensionless, while the second term has a unit of length. Furthermore, in Eq. (\ref{eq:frk_std}), we know Block matrix $\boldsymbol{J}_{a1}$ is dimensionless and $ \boldsymbol{J}_{a2} $ has a unit of $\frac{\text{1}}{\text{length}}$. As a result, we conclude that the Jacobian for this particular group of manipulators is dimensionless.

   \textit{Case 2: PMs with rotational actuators.} For PMs with rotational actuators $\dot{\boldsymbol{q}}_a$ has unit of $\frac{\text{angle}}{\text{time}}$ while the unit of $\boldsymbol{V}_p$ is unchanged. Furthermore, the matrix $\boldsymbol{J}_{a1}$ has a unit of length and $\boldsymbol{J}_{a2}$ is dimensionless for this group of PMs. Consequently, the resulting Jacobian $\boldsymbol{J}_{dh}$ to has a unit of length which is consistent.

   Because entries of $\boldsymbol{J}_{dh}$ are either dimensionless or dimensionally homogeneous, its condition number or singular values have physical significance and can be used to measure the dexterity of the manipulator.

    The next section demonstrates how to derive it by considering a relevant example: a four DoF (degrees of freedom) $T_yT_zR_xR_y$ type Parallel Manipulator (PM).

   \section{Example}

\begin{figure}[htb!]
   	\centering
   	\includegraphics[width=0.47\textwidth]{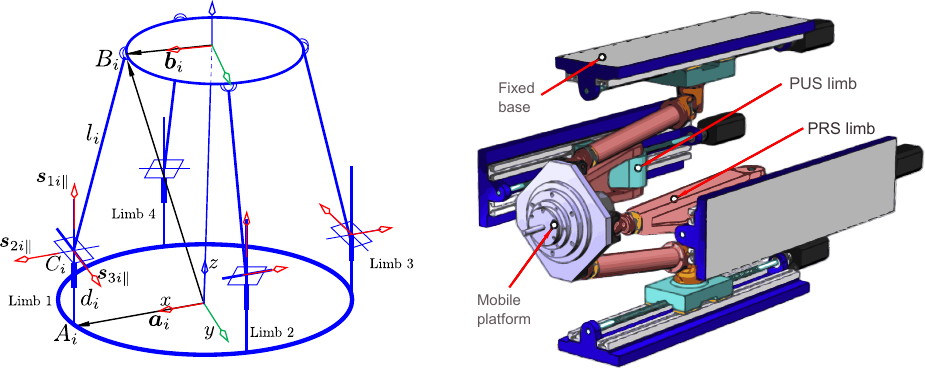}
   	\caption{Four DoF PM}
   	\label{fig:2rrru/rrs}
   \end{figure}

   The mechanism depicted in Fig. \ref{fig:2rrru/rrs} is a $T_y T_z R_x R_y$ type 4 DoF PM \cite{Li2013} with a PUS joint order in the first and third limbs, and a PRS type joint sequence in the second and fourth limbs. The P joint is parallel to the $z$-axis, while the R joint in the PRS limb is parallel to the $x$-axis, and the U joint in the PUS limb has axes parallel to the $x$ and $y$ axes, respectively. The mechanism is capable of rotating about the $x$ and $y$ axes, as well as translating along the $y$ and $z$ directions. However, due to the presence of revolute joints in the second and fourth limbs, the mechanism is constrained in terms of $x$-axis translation and $z$-axis rotations, making it a zero-torsion type PM mechanism.
   The DoF of the mechanism can also be determined by employing Tsai's DoF formula, which is expressed as follows:    
   \begin{equation}
   	\begin{aligned}
   		F &= \lambda(n-j-1) + \sum_{i=1}^{j} f_i   \\ 4&= 6(10-12-1)+22  	\label{eq:tsai_frm}
    \end{aligned}
   \end{equation}    
   Point $A_i$ at the based is location of limbs while $B_i$ is the center of spherical joints. Point $C_i$ is the center universal joints for the first and third limbs while it is the center of revolute joints for limbs 2 and 4. The position vector $\boldsymbol{a}_i$ is extended from origin moving frame to the $i^{th}$ spherical joint while $\boldsymbol{b}_i$ is extended from the fixed frame to the point $A_i$. The direction vector $\boldsymbol{s}_{ji\parallel}$ is associated to each joint axis.

	To appropriately represent the motion of the moving plate, we need four points and for convenience these points are chosen to be the center of the spherical joints. Expanding Eq. (\ref{eq:pts_vel}) to the four points located at the center of spherical joints at the moving plate, we get a $ 12 \times 6$ matrix that relates the $i^{th}$ points linear velocity to the moving plate center velocity ($\boldsymbol{\dot{\mathscr{x}}}$) as

	\begin{equation}
			\begin{bmatrix} v_{1x} \\ v_{1y} \\ v_{1z} \\ \vdots \\  v_{4x} \\ v_{4y} \\ v_{4z}\end{bmatrix} =       \begin{bmatrix}
			1 & 0 & 0 & 0 & a_{1z} & -a_{1y} \\
			0 & 1 & 0 & -a_{1z} & 0 & a_{1x} \\
			0 & 0 & 1 & a_{1y} & -a_{1x} & 0 \\
			\vdots & \vdots & \vdots & \vdots & \vdots & \vdots  \\
			1 & 0 & 0 & 0 & a_{4z} & -a_{4y} \\
			0 & 1 & 0 & -a_{4z} & 0 & a_{4x} \\
			0 & 0 & 1 & a_{4y} & -a_{4x} & 0
		\end{bmatrix}  \begin{bmatrix} v_x \\ v_y \\ v_z \\ \omega_x \\ \omega_y \\ \omega_z \end{bmatrix} \label{eq:v_gen}
	\end{equation}

	 The points' linear velocity in Eq. (\ref{eq:v_gen}) includes 12 components, three for each point. Therefore, we have many dependent motions, yet we only require four components. Referring to Eq. (\ref{eq:vi_comp}), none of the components encompass the desired motion of the moving plate. Hence, we need to formulate a selection matrix that can combine components from different points and obtain a nominal velocity that describes the motion of the moving plate. As the independent motion of the moving plate for this manipulator are $v_y, v_z, \omega_x$ and $\omega_y$, combining $v_{iy}$ and $v_{iz}$ components can sufficiently describe the manipulator's motion. However, a combination of component is not unique and one can freely choose one of the following pairs.
		\begin{equation}
			\begin{aligned}
				\text{Limb 1:} ~&~ (v_{1y}, v_{2z}), (v_{1y}, v_{3z}), (v_{1y},v_{4z}) \\
				\text{Limb 2:} ~&~ (v_{2y}, v_{1z}), (v_{2y}, v_{3z}), (v_{2y},v_{4z}) \\
				\text{Limb 3:} ~&~ (v_{3y}, v_{1z}), (v_{3y}, v_{2z}), (v_{3y},v_{4z}) \\
				\text{Limb 4:} ~&~ (v_{4y}, v_{1z}), (v_{4y}, v_{2z}), (v_{4y},v_{3z}) \\
			\end{aligned} \label{eq:list}
		\end{equation}
	For this particularly case, we selected the following combination from Eq. (\ref{eq:list}).

	\begin{equation}
		\begin{aligned}
			\text{Limb 1:} ~&~ (v_{1y}, v_{2z})  \\
			\text{Limb 2:} ~&~ (v_{2y}, v_{3z})  \\
			\text{Limb 3:} ~&~ (v_{3y},v_{4z})   \\
			\text{Limb 4:} ~&~ (v_{4y}, v_{1z})  \\
		\end{aligned} \label{eq:list_selected}
	\end{equation}

	By utilizing Eq. (\ref{eq:list_selected}), we can establish the extended selection matrix as

	\begin{equation}
		\begin{aligned}[b]
			\boldsymbol{S}
			&= \left[
			\begin{matrix}
					0 & -\dfrac{a_{2x}}{a_{1x} - a_{2x}} & 1 & 0 & \dfrac{a_{1x}}{a_{1x} - a_{2x}}  & 0 \\
					0 & 0 								 & 0 & 0 & -\dfrac{a_{3x}}{a_{2x} - a_{3x}} & 1 \\
					0 & 0 								 & 0 & 0 & 0 & 0 \\
					0 & -\dfrac{a_{4x}}{a_{1x} - a_{4x}} & 1 & 0 & 0 & 0 \\
			\end{matrix}
			\right.  \   \\
			&\qquad   \ \left.
			\begin{matrix}
			  0 & 0 & 0 & 0 & 0 & 0 \\
			  0 & \dfrac{a_{2x}}{a_{2x} - a_{3x}} & 0 & 0 & 0 & 0 \\
			  0 & -\dfrac{a_{4x}}{a_{3x} - a_{4x}} & 1 & 0 & \dfrac{a_{3x}}{a_{3x} - a_{4x}} & 0 \\
			  0 & 0 & 0 & 0 & \dfrac{a_{1x}}{a_{1x} - a_{4x}} & 0
			\end{matrix}
			\right]
		\end{aligned} \label{eq:sel_mat}
	\end{equation}
	It is quite important to note that $\boldsymbol{S}$ in Eq. (\ref{eq:sel_mat}) is dimensionless and same as to that the usual selection matrix in terms of units. However, the usual selection matrices \cite{Nigatu2023} have entries of 1s and 0s unlike the extended selection matrix derived in the paper.  	
	Then, multiplying Eq. (\ref{eq:sel_mat}) with Eq. (\ref{eq:v_gen}),  matrix$\boldsymbol{V}_{ps}  \in {\mathbb{R}}^{4 \times 4}$ relates the nominal velocity ($\boldsymbol{v}_{ps}$) and independent Cartesian velocity of the moving plate as in Eq. (\ref{eq:nom_vel}).

	\begin{equation}
		\begin{aligned}[b]
			\begin{bmatrix} \underline{v_1} \\ \underline{v_2} \\ \underline{v_3} \\ \underline{v_4} \end{bmatrix}
			&= \left[
			\begin{matrix}
				\dfrac{a_{1x}}{a_{1x} - a_{2x}} - \dfrac{a_{2x}}{a_{1x} - a_{2x}} & 1\\
				\dfrac{a_{2x}}{a_{2x} - a_{3x}} - \dfrac{a_{3x}}{a_{2x} - a_{3x}} & 1  \\
				\dfrac{a_{3x}}{a_{3x} - a_{4x}} - \dfrac{a_{4x}}{a_{3x} - a_{4x}} & 1  \\
				\dfrac{a_{1x}}{a_{1x} - a_{4x}} - \dfrac{a_{4x}}{a_{1x} - a_{4x}} & 1  \\
			\end{matrix}
			\right.  \  \\
			&\qquad   \ \left.
			\begin{matrix}
				\dfrac{a_{1y}a_{1x} - a_{2x}a_{1z} + a_{1x}a_{3z}}{a_{1x} - a_{2x}} & -a_{1x}  \\
				\dfrac{a_{3y}a_{2x} + a_{2x}a_{3z} - a_{3x}a_{3z}}{a_{2x} - a_{3x}} & -a_{2x}  \\
				\dfrac{a_{3y}a_{3x} + a_{3x}a_{4z} - a_{4x}a_{3z}}{a_{3x} - a_{4x}} & -a_{3x} \\
				\dfrac{a_{1y}a_{1x} + a_{1x}a_{4z} - a_{4x}a_{1z}}{a_{1x} - a_{4x}} & -a_{1x}
			\end{matrix}
			\right]
		\end{aligned} 	\begin{bmatrix}
		vy \\
		vz \\
		wx \\
		wy
	\end{bmatrix} \label{eq:nom_vel}
	\end{equation}

	where, $\begin{bmatrix} \underline{v_1} \\ \underline{v_2} \\ \underline{v_3} \\ \underline{v_4} \end{bmatrix} =  \begin{bmatrix}
		\dfrac{(v_{2y} + v_{1z})a_{1x} - (v_{1y} + v_{1z})a_{2x}}{a_{1x} - a_{2x}} \\
		\dfrac{(v_{3y} + v_{2z})a_{2x} - (v_{2y} + v_{2z})a_{3x}}{a_{2x} - a_{3x}} \\
		\dfrac{(v_{4y} + v_{3z})a_{3x} - (v_{3y} + v_{3z})a_{4x}}{a_{3x} - a_{4x}} \\
		\dfrac{(v_{4y} + v_{1z})a_{1x} - (v_{1y} + v_{1z})a_{4x}}{a_{1x} - a_{4x}} \\
	\end{bmatrix}$

	The inverse Jacobian of the manipulator is obtained through the analytic screw theory method and is given as shown in Eq. (\ref{eq:irk}). The first four row of $\boldsymbol{G}^T$ represents the motion Jacobian while the last two depicts the structural constraints. Hence, $\boldsymbol{G}_c^T\boldsymbol{\dot{\mathscr{x}}} =\boldsymbol{0}$ is always satisfied.  

	\begin{equation}
		\dot{\boldsymbol{q}} = \begin{bmatrix} \boldsymbol{G}_a^T \\ \boldsymbol{G}_c^T \end{bmatrix}\boldsymbol{\dot{\mathscr{x}}}  = \begin{bmatrix}
			\dfrac{\boldsymbol{n}_{1}^T}{\boldsymbol{n}_{1}^T\boldsymbol{s}_{11\parallel}} &  \dfrac{(\boldsymbol{n}_{1} \times \boldsymbol{a}_1)^T}{\boldsymbol{n}_{1}^T\boldsymbol{s}_{11\parallel}} \\
			\dfrac{\boldsymbol{l}_{2}^T}{\boldsymbol{l}_{2}^T\boldsymbol{s}_{12\parallel}} &  \dfrac{(\boldsymbol{l}_{2} \times \boldsymbol{a}_2)^T}{\boldsymbol{l}_{2}^T\boldsymbol{s}_{12\parallel}} \\
			\dfrac{\boldsymbol{n}_{3}^T}{\boldsymbol{n}_{3}^T\boldsymbol{s}_{13\parallel}} &  \dfrac{(\boldsymbol{n}_{3} \times \boldsymbol{a}_3)^T}{\boldsymbol{n}_{3}^T\boldsymbol{s}_{13\parallel}} \\
			\dfrac{\boldsymbol{l}_{4}^T}{\boldsymbol{l}_{4}^T\boldsymbol{s}_{14\parallel}} &  \dfrac{(\boldsymbol{l}_{4} \times \boldsymbol{a}_4)^T}{\boldsymbol{l}_{4}^T\boldsymbol{s}_{14\parallel}} \\
			\boldsymbol{s}_{22\parallel}^T &  (\boldsymbol{s}_{22\parallel} \times \boldsymbol{a}_2)^T  \\
			\boldsymbol{s}_{24\parallel}^T &  (\boldsymbol{s}_{24\parallel} \times \boldsymbol{a}_4)^T  \\  \end{bmatrix}  \begin{bmatrix} \boldsymbol{v} \\ \boldsymbol{\omega}\end{bmatrix}  \label{eq:irk}
	\end{equation}
	where $  \boldsymbol{G}_a^T \in   {\mathbb{R}}^{4 \times 6}  $   and  $\boldsymbol{G}_c^T \in {\mathbb{R}}^{2 \times 6}  $ and $\boldsymbol{n}_i = \boldsymbol{s}_{3i\parallel} \times \boldsymbol{s}_{2i\parallel}$. $\boldsymbol{l}_i $ is a vector extending from $C_i$ to $B_i$.

	The first term of $\boldsymbol{G}^T$ is dimensionless while the second term has a unit of length. Hence, units of the  inverse Jacobian of this manipulator is inconsistent and must be changed to dimensionless or consistent unit.

	The forward Jacobian $\boldsymbol{J}_a \in {\mathbb{R}}^{6 \times 3}$ is analytically obtained by inverting $\boldsymbol{G}^{-T}$ as in Eq. (\ref{eq:Ja}).

	\begin{equation}
		\boldsymbol{J}_a=
		\!\begin{aligned}
			&
			\left[\begin{matrix}
				\boldsymbol{G}_{av}^{-T}\boldsymbol{G}_{aw}^T(\boldsymbol{G}_{cw}^T-\boldsymbol{G}_{cv}^T\boldsymbol{G}_{av}^{-T} \times\\
				-(\boldsymbol{G}_{cw}^T - \boldsymbol{G}_{cv}^T\boldsymbol{G}_{av}^{-T}\boldsymbol{G}_{aw})^{-1} \times
			\end{matrix}\right.\\
			&\qquad\qquad
			\left.\begin{matrix}
				{} \boldsymbol{G}_{av}^{-T}+ \boldsymbol{G}_{aw}^T)^{-1}\boldsymbol{G}_{cv}^T\boldsymbol{G}_{av}^{-T}\\
				{} \boldsymbol{G}_{cv}^T\boldsymbol{G}_{av}^{-T}
			\end{matrix}\right]
		\end{aligned}  \label{eq:Ja}
	\end{equation}

	By substituting Eq. (\ref{eq:sel_mat}) and Eq. (\ref{eq:v_gen}) into Eq. (\ref{eq:vps}), and subsequently replacing $\boldsymbol{\dot{\mathscr{x}}}$ with $\boldsymbol{J}_a\dot{\boldsymbol{q}}$ in Eq. (\ref{eq:vps}), we derive the $4 \times 4$ dimensionless Jacobian as discussed in \textit{case 1}.

	\subsection{Numerical Evaluation}

In order to verify the correctness of the derived dimensionally homogeneous Jacobian, the distribution of the condition number ($k$) for the manipulator over the entire workspace is evaluated using geometric and motion parameters outlined in Table \ref{table:param}.

\begin{table}[htb!]
	\caption{Structural and pose parameters of the PM.}
	\centering
	\begin{tabular}{|c|c|}
		\hline
		Parameters& value \\
		\hline
		Radius of the moving plate $(r_a)$ & 200~mm \\
		\hline
		Radius of the base plate $(r_b)$ & 450~mm \\
		\hline
		Fixed length link $(l)$ & 687~mm \\
		\hline
		$\theta$ & $\pm 50 ^o$\\
		\hline
		$\psi$ & $\pm 50^o$ \\
		\hline
		$y$ & 0~mm \\
		\hline
		$z$ & 100-200~mm   \\
		\hline
	\end{tabular} \label{table:param}
\end{table}

It is known that in parallel manipulator design, the condition number ($k$) of the Jacobian matrix can be used as a performance measure to evaluate the quality of motion, precision, and stability of the manipulator. The best value of $k$ is 1 which is the minimum possible value and it indicates that all columns (or rows) of the Jacobian matrix matrix are orthogonal to each other. This implies that the system of equations is well-conditioned and the solution will not be overly sensitive to errors in the data or to small changes in the input. This can be interpreted as that the manipulator is \textit{isotropic} \cite{Merlet2007}. As $k$ increases beyond 1, the system of equations becomes increasingly ill-conditioned. This means that the solution may be very sensitive to errors in the data or to small changes in the input and hence the manipulator is approaching to singularity. Contrary to this, if $k$ values is small enough or remains closer to 1, it can be interpreted the manipulator is away from singular configuration.

\begin{figure}[htb!]
	\centering
	\includegraphics[width=0.5\textwidth]{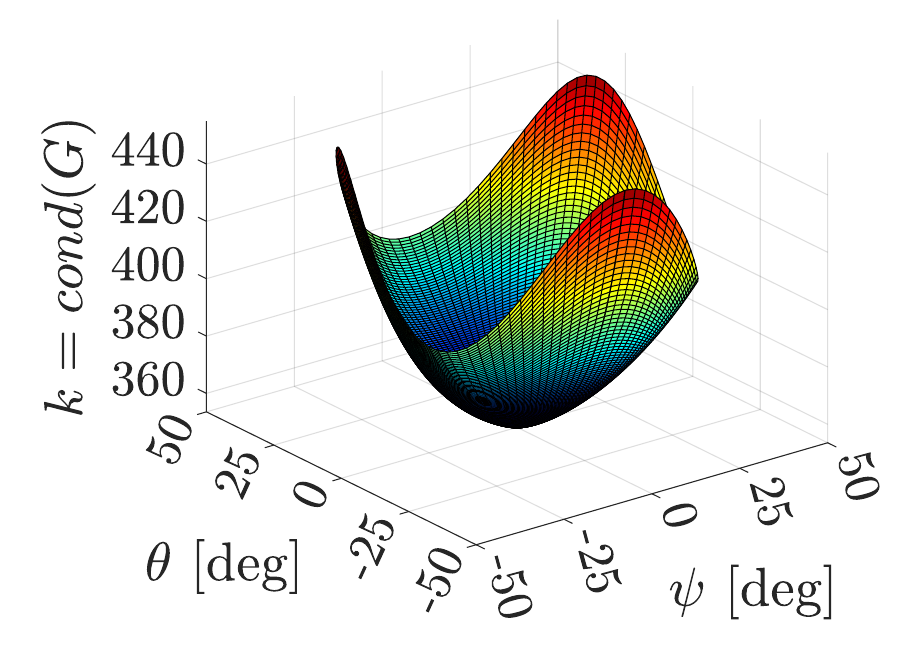}
	\caption{$k=cond(\boldsymbol{G})$}
	\label{fig:condG}
\end{figure}

\begin{figure}[htb!]
	\centering
	\includegraphics[width=0.5\textwidth]{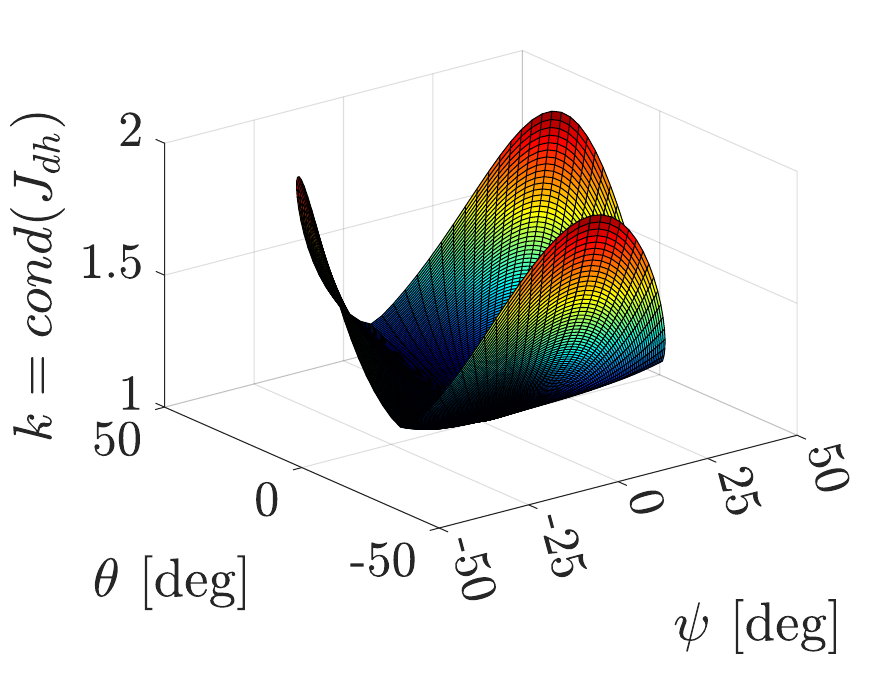}
	\caption{$k=cond(\boldsymbol{J}_{dh})$}
	\label{fig:condjdh}
\end{figure}

Accordingly, $k=cond(\boldsymbol{G}^T)$ is first computed and the result is shown in Fig. \ref{fig:condG} over the rotational workspace. The simulation result indicate a substantial increase in the condition number, which do not to adequately reflect the physical properties of the manipulator. Consequently, $cond(\boldsymbol{J}_{dh})$ was determined, as shown in Fig. \ref{fig:condjdh}. In the rotational workspace, the value of $k$ remained low and near 1. This value of $k$ can properly indicate whether the manipulator is far or approaching a singular configuration.

\begin{figure}[htb!]
	\centering
	\includegraphics[width=0.5\textwidth]{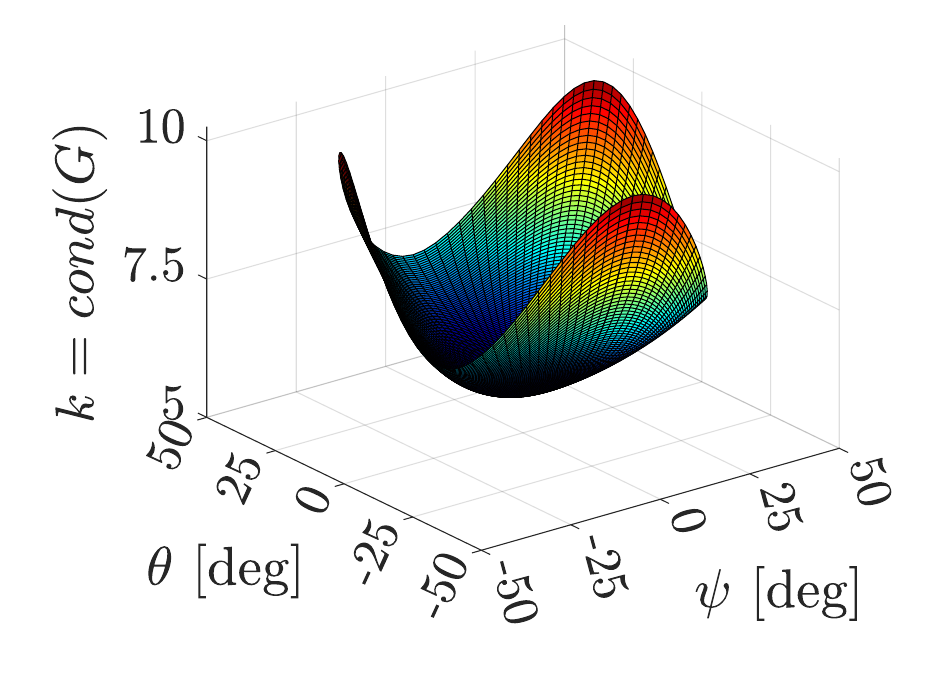}
	\caption{$k=cond(\boldsymbol{J}_{dh})$}
	\label{fig:condG_mt}
\end{figure}

Additionally, the sensitivity of $k$ of both Jacobians to unit changes is evaluated over the rotational workspace. The results shows significant discrepancy in the condition number of the conventional Jacobian when units are changed from millimeter to meters, as depicted in Fig. \ref{fig:condG_mt}. Comparing this result with Fig. \ref{fig:condG} can be considered more optimized even-though nothing is changed but unit. However, the value of $cond(\boldsymbol{J}_{dh})$, when measured in meters, remained unchanged. The result of $cond(\boldsymbol{J}_{dh})$ in meter is not provided here because it is the same as to that of shown in Fig. \ref{fig:condjdh}. Hence, $cond(\boldsymbol{J}_{dh})$ is invariant under the change of units.
As previously mentioned, the choice of component combinations is not unique. Hence, we have the flexibility to choose various pairs of $v_{iy}$ and $v_{iz}$ from the provided candidates. For instance, by selecting $(v_{1y}, v_{3z}), (v_{2y}, v_{4z}), (v_{3y}, v_{1z})$, and $(v_{4y}, v_{2z})$, we can derive the following selection matrix.

\begin{equation*}
	\begin{aligned}[b]
		\boldsymbol{S}
		&= \left[
		\begin{matrix}
			0 & -\dfrac{a_{2x}}{a_{3x} - a_{2x}} & 1 & 0 & \dfrac{a_{3x}}{a_{3x} - a_{2x}}  & 0 \\
			0 & 0          & 0 & 0 & -\dfrac{a_{4x}}{a_{2x} - a_{4x}} & 1 \\
			0 & 0          & 0 & 0 & 0 & 0 \\
			0 & -\dfrac{a_{4x}}{a_{2x} - a_{4x}} & 1 & 0 & 0 & 0 \\
		\end{matrix}
		\right.  \   \\
		&\qquad   \ \left.
		\begin{matrix}
			0 & 0 & 0 & 0 & 0 & 0 \\
			0 & \dfrac{a_{2x}}{a_{2x} - a_{4x}} & 0 & 0 & 0 & 0 \\
			0 & -\dfrac{a_{1x}}{a_{3x} - a_{1x}} & 1 & 0 & \dfrac{a_{3x}}{a_{3x} - a_{1x}} & 0 \\
			0 & 0 & 0 & 0 & \dfrac{a_{2x}}{a_{2x} - a_{4x}} & 0
		\end{matrix}
		\right]
	\end{aligned} \label{eq:sel_mat2}
\end{equation*}

Then, with this selection matrix, we establish the dimensionally homogeneous Jacobian as shown in Eq. (\ref{eq:jdh}) and the condition number distribution over the workspace is evaluated. The simulation has shown the same result to that of the dimensionally homogeneous Jacobian obtained using Eq. (\ref{eq:sel_mat}).  
  
This consistent property of the Jacobian matrix is quite important while using the condition number as a measure of performance or computing the dexterity for parameter optimization of PMs.

\section{Conclusion}
This paper introduces an extended selection matrix to formulate a point-based, dimensionally homogeneous Jacobian of various constrained parallel manipulators. The proposed method allows for the derived Jacobian's condition number and singular values to be utilized as a performance index and optimization with unit independence.  

To validate the proposed approach, the condition number ($k$) for both the conventional Jacobian ($\boldsymbol{G}$) and the dimensionally homogeneous Jacobian ($\boldsymbol{J}_{dh}$) across the rotational workspace were compared. Simulation results indicated a large value of $k$ for $\boldsymbol{G}$ and a remarkably stable value of $k$ for $\boldsymbol{J}_{dh}$.

Further, we reassessed the distribution of the $k$ value for the two Jacobians by changing the units from millimeters to meters. The results confirmed that $k$ of $\boldsymbol{G}$ varied significantly, while $k$ of $\boldsymbol{J}_{dh}$ remained consistent, irrespective of the unit change. This phenomenon proves the dimensional homogeneity of the proposed Jacobian, where both the linear and angular parts exhibit similar value distributions and are not unit-dependent. As a result, our method allows for the correct optimization of the manipulators with mixed DoFs.
By employing the proposed approach for different manipulators with mixed DoFs, we can confidently assess and optimize their performance.

\section*{ACKNOWLEDGMENT}

This work was supported by Korea Institute of Science and Technology (KIST), under Grant 2E32302.

\bibliographystyle{IEEEtran}
\bibliography{bib_file_iccas}
%
%
%
%
%
%
%
%

\end{document}